\documentclass{article}

\usepackage[english]{babel}
\usepackage[letterpaper,top=2cm,bottom=2cm,left=3cm,right=3cm]{geometry}

\usepackage{amsmath}
\usepackage{graphicx}
\usepackage{booktabs}
\usepackage{authblk}

\usepackage[hidelinks]{hyperref}

\title{Visually grounded emotion regulation via diffusion models and user-driven reappraisal}

\author[1,2]{Edoardo Pinzuti\thanks{\href{mailto:Edoardo.Pinzuti@lir-mainz.de}{Edoardo.Pinzuti@lir-mainz.de}}}
\author[2,3]{Oliver Tüscher\thanks{\href{mailto:Oliver.Tuescher@uk-halle.de}{Oliver.Tuescher@uk-halle.de}}}
\author[4]{André Ferreira Castro\thanks{\href{mailto:andre.ferreira.castro@tum.de}{Andre.Ferreira.Castro@tum.de}}}

\affil[1]{Leibniz Institute for Resilience Research, Mainz, Germany}
\affil[2]{University Medicine Halle (Saale) of the Martin Luther University Halle-Wittenberg (MLU), Halle (Saale), Germany}
\affil[3]{German Center for Mental Health (DZPG), partner site Halle-Jena-Magdeburg, Halle (Saale), Germany}
\affil[4]{School of Life Sciences, Technical University of Munich, Freising 85354, Germany}

\date{}

\begin{document}
\maketitle

\begin{abstract}
Cognitive reappraisal is a key strategy in emotion regulation, involving reinterpretation of emotionally charged stimuli to alter affective responses. Despite its central role in clinical and cognitive science, real-world reappraisal interventions remain cognitively demanding, abstract, and primarily verbal in nature. This reliance on higher-order cognitive and linguistic processes can be especially impaired in individuals with trauma, depression, or dissociative symptoms, limiting the effectiveness of standard approaches. Here, we propose a novel, visually based augmentation of cognitive reappraisal by integrating large-scale text-to-image diffusion models into the emotional regulation process. Specifically, we introduce a system wherein users reinterpret emotionally negative images via spoken reappraisals, which are then transformed into supportive, emotionally congruent visualizations using stable diffusion models with a fine-tuned IP-adapter module. This generative transformation visually instantiates users’ cognitive reappraisals while maintaining structural similarity to the original stimuli, thus externalizing and reinforcing regulatory intent. To evaluate this approach, we conducted a within-subjects experiment ($N = 20$) using a modified cognitive emotion regulation (CER) task. Participants reappraised or described aversive images from the international affective picture system (IAPS), with or without AI-generated visual feedback. Results indicate that AI-assisted reappraisal significantly reduced negative affect relative to both non-AI reappraisal and control conditions. Further analyses show that sentiment alignment between participant reappraisals and generated images correlates with affective relief, suggesting that multimodal coherence enhances regulatory efficacy. Our findings highlight the feasibility of using generative visual support for cognitive reappraisal. This work opens a new interdisciplinary direction at the intersection of generative AI, affective computing, and therapeutic technology design.
\end{abstract}

\section{Introduction}

Emotion regulation is a central process in human psychological functioning, underpinning a wide range of outcomes related to mental health, interpersonal behavior, and adaptive decision-making \cite{gross2002emotion,lincoln2022role,gross2015emotion}. Within this domain, cognitive reappraisal—defined as the reinterpretation of an emotional stimulus to modify its impact—has emerged as a particularly effective strategy \cite{hajcak2010emotion,riepenhausen2022positive}. Empirical research has consistently demonstrated that reappraisal can reduce negative affect, lower physiological arousal, and serve as a protective factor against stress-related disorders across both clinical and non-clinical populations \cite{han2023psychological,in2021effects, southward2022more}. It is also a cornerstone of cognitive-behavioral therapy (CBT), supported by a robust body of experimental and neuroimaging data \cite{bryant2021reappraisal,buhle2014cognitive, halfmann2021neural}. These findings have contributed to its widespread use in both laboratory paradigms and therapeutic interventions, underscoring its relevance as a tool for promoting emotional resilience \cite{mcrae2012unpacking, montfort2024role}.

Standardized laboratory procedures, such as the Cognitive Emotion Regulation (CER) task, have provided key insights into the mechanisms of reappraisal \cite{kanske2011regulate}. In such paradigms, participants are typically instructed to mentally reinterpret negatively valenced stimuli—often drawn from the International Affective Picture System (IAPS, \cite{lang1997international})—in ways that diminish their emotional salience \cite{erk2010acute, sandner2021cognitive}. Functional neuroimaging studies employing these tasks have shown that successful reappraisal is associated with reduced amygdala activity \cite{banks2007amygdala, klumpp2017predicting} and enhanced activation in prefrontal regions implicated in cognitive control \cite{ray2012anatomical, morawetz2016changes}. However, despite its empirical validity, cognitive reappraisal remains a cognitively demanding and introspective technique that hinges on abstract verbal reasoning \cite{vy2024relationship, sass2013emotional, mcrae2016cognitive}. This requirement places significant demands on executive function, which can limit the strategy's effectiveness in real-world contexts, particularly during emotionally intense episodes or among individuals with cognitive impairments or those suffering from mental health conditions such as depression, anxiety disorders and post-traumatic stress disorder (PTSD) \cite{gross1995emotion,kraaij2019behavioral, jacob2016study, lincoln2022role}.

Given these constraints, there is increasing interest in augmenting reappraisal with computational tools that can reduce cognitive load and facilitate more accessible emotion regulation \cite{bucci2019digital,mitsea2023digitally}. Despite the growing interest in applying artificial intelligence (AI) to mental health support, the integration of generative AI, such as, large language models (LLMs), into emotion regulation interventions remains nascent \cite{dehbozorgi2025application, miner2017talking, rubin2025comparing, pataranutaporn2023influencing,glickman2025human, sharma2023human}. Recent developments in conversational agents powered by LLMs have enabled more fluent and empathetic interactions, with some studies demonstrating the capacity of LLMs to generate high-quality cognitive reappraisals \cite{zhan2024reappraisal, chiu2024computational}. However, these systems are largely limited to passive, text-based outputs that do not actively engage users in multimodal processing \cite{prochaska2021therapeutic, vaidyam2019chatbots}. 

Meanwhile, advances in text-to-image generation—particularly through latent diffusion models such as stable diffusion—have enabled efficient and high-quality synthesis of photorealistic and semantically rich images from textual prompts \cite{blattmann2023stable,zhang2023adding, podell2023sdxlimprovinglatentdiffusion}. More recently, image-conditioned variants like IP-Adapter architectures have extended this capability by allowing models to receive both reference images and guiding text, preserving visual structure while modifying semantic or emotional content \cite{xu2023ipadapter}. Early explorations into AI-generated imagery for emotional reflection have shown promise in enhancing affective engagement, yet they remain disconnected from explicit reappraisal strategies and lack systematic evaluation of their emotional impact \cite{yang2024emogen,lomas2024improved}. Despite their expressive potential, these models have not yet been applied to affective computing or emotion regulation due potential societal representational biases in those models \cite{luccioni2023stable}. Existing applications have largely focused on aesthetic style transfer, avatar personalization, or facial expression synthesis, without grounding the visual output in user-driven cognitive reinterpretation \cite{mittenentzwei2024ai}. Thus, the potential of image-conditioned generative models to serve as interactive, affective scaffolds—instantiating user-authored reappraisals in visually recontextualized form—remains unexplored \cite{ruppert2018positive, yamada2018exploratory, pearson2015mental}. The opportunity to leverage these technologies for real-time co-regulation of emotional meaning between human users and AI agents represents a critical, unmet need in both computational psychiatry and human-computer interaction \cite{bell2024advances,blackwell2021mental}.

To address this gap, we introduce a novel framework for visual cognitive reappraisal that integrates generative diffusion models into an emotion regulation task. Participants reinterpret negative images through verbal reappraisal, which is transcribed and used—alongside the original image—to condition a stable diffusion XL model with IP-Adapter. The resulting AI-generated image reflects the user’s reinterpretation, offering a visually grounded, emotionally resonant scaffold for regulation. In a controlled study ($N = 20$), we show that this AI-assisted approach significantly reduces negative affect compared to traditional mental-only reappraisal. Furthermore, affective improvement was predicted by sentiment alignment between the reappraisal prompt and the generated image, highlighting the role of multimodal coherence in emotional regulation. Our findings highlight the value of aligning visual output with user intent, and demonstrate how generative AI can act as a real-time cognitive amplifier, bridging abstract thought and emotional transformation. This work opens a new interdisciplinary direction at the intersection of generative AI, affective computing, and therapeutic technology design.

\section{Methods}
\label{Methods}
\subsection{Subjects}
Twenty healthy adults (10 female, 10 male; age range: 25–55 years, mean $= 35.45 \pm 8.86$) participated in the study. Participants were screened for a history of psychiatric or neurological disorders. Exclusion criteria included current use of psychotropic medication and insufficient fluency in one of the four supported languages (English, Italian, German, French). Participants provided informed consent in accordance with institutional ethical guidelines. Language preference was respected during task interaction: participants could complete the task in their native language or in English if fluent. Three participants opted for English over their native language. Verbal responses were recorded in the preferred language and subsequently translated to English for model input and analysis to align with model training distributions and avoid linguistic bias.

\subsection{Experimental design}
\label{Experimental design}
We used a modified Cognitive Emotion Regulation (CER) task \cite{kanske2011regulate} to compare traditional and AI-assisted reappraisal. The CER task is a widely adopted paradigm for studying cognitive strategies that alter affective experience \cite{erk2010acute, sandner2021cognitive}. We adapted it to incorporate real-time generative image feedback. Each participant completed trials under one of four within-subject conditions in a $2 \times 2$ design: Describe, Describe AI, Reappraise, and Reappraise AI.

At the start of each trial, participants viewed a neutral or negative image from the International Affective Picture System (IAPS; \cite{lang1997international}) for 4 seconds. This period was intended to allow sufficient time to visually encode the stimulus. Afterward, a microphone icon signaled participants to speak aloud, either by describing the image (Describe conditions) or by reinterpreting it in a more positive light (Reappraise conditions). The original image remained onscreen during this verbalization period to preserve visual context. Spoken input was chosen over typing to reduce distraction and cognitive switching. Participants were given 12 seconds to speak.

In the \textbf{AI conditions}, verbal utterances were transcribed in real time using OpenAI’s Whisper-turbo speech recognition system \cite{radford2022robust}. If spoken in a language other than English, transcripts were automatically translated without manual correction. These English transcripts, together with the original image, were passed as input to a Stable Diffusion XL model equipped with an IP-Adapter module \cite{xu2023ipadapter}. To account for image generation time, a gray screen was presented for a duration of 4 seconds. The resulting image reflected the participant’s interpretation and was shown for 3 seconds.
To ensure timing was matched across all conditions, a gray screen was also shown for 4 seconds after
the verbal phase in the \textbf{non-AI conditions}.
After this, participants rated their emotional state using a visual analog scale ranging from very positive to very negative. The rating interface consisted of a horizontal slider positioned below five emoticon anchors representing discrete levels of valence. The slider was mapped to integer values from 1 to 9. Subsequently, for visualization and analysis, we remapped affective rating scale such that 0 indicated a neutral response, with values ranging from –2 (strongly negative) to +2 (strongly positive). Participants were trained on how to interpret each emoticon prior to beginning the task.

Training also included examples for both verbal conditions. In the Reappraise conditions, participants were instructed to create an alternative, positive narrative or outcome for the scene (e.g., “This person will recover,” or “They are being rescued”). Describe trials served as a low-level verbal baseline, allowing comparisons that isolate the contribution of reinterpretation and/or generative visual augmentation.

\subsection{Stable diffusion model XL}

Stable Diffusion XL (SDXL) is a high-capacity text-to-image generative model that extends the latent diffusion framework introduced in prior work~\cite{podell2023sdxlimprovinglatentdiffusion}. Rather than generating images directly in pixel space, SDXL operates in a compressed latent space defined by a pretrained variational autoencoder (VAE), enabling efficient sampling while preserving high-resolution image quality. The core generative process is implemented via a U-Net denoising architecture, conditioned on text embeddings produced by a frozen CLIP-based encoder~\cite{che2023enhancing}. This encoder converts natural language prompts into a sequence of token representations, which are injected into the U-Net through a series of cross-attention layers, guiding image synthesis at multiple spatial scales. Compared to earlier iterations of Stable Diffusion, SDXL introduces a dual-text encoder configuration and supports high-fidelity generation at resolutions up to $1024 \times 1024$ pixels. These improvements yield more robust prompt interpretability, as well as finer control over stylistic and semantic content.

\subsection{IP-Adapter for Visual Conditioning}

To enable joint conditioning on both visual and textual information, we incorporate the IP-Adapter architecture proposed by Ye et al. \cite{xu2023ipadapter}, adapting it to function within a frozen SDXL backbone. The IP-Adapter is a lightweight plug-in module (approximately 22 million parameters) designed to inject reference image features into the diffusion process via a decoupled cross-attention mechanism, without modifying the pre-trained model weights.

A frozen CLIP ViT-H/14 encoder is used to extract a global embedding from a reference image \cite{radford2021learning}. This embedding is projected through a small trainable network—comprising a linear layer followed by LayerNorm—into a fixed-length sequence of image tokens (see section ~\ref{Dataset construction for stable diffusion IP-adapter}). These tokens share the same dimensionality as the text embeddings from SDXL’s CLIP-based text encoder, allowing direct integration into the U-Net's attention layers.

In each cross-attention block of the SDXL U-Net, the IP-Adapter introduces a parallel attention stream dedicated to image-based features. The query projection matrix $W^Q$ is shared between modalities, while separate key and value projections ($W^{K'}$, $W^{V'}$) are learned for the image pathway. The text-based and image-based outputs, denoted $O_t$ and $O_i$ respectively, are initially combined via simple addition:
\begin{equation}
O_{\text{combined}} = O_t + O_i
\end{equation}
To allow for finer control over the contribution of the image features, we introduce a conditioning scale parameter $\lambda \in [0, 1]$:
\begin{equation}
O_{\text{combined}} = O_t + \lambda \cdot O_i
\end{equation}
Here, $\lambda$ modulates the relative influence of visual guidance. A lower value of $\lambda$ emphasizes the text prompt, while higher values strengthen the impact of the image features. We adopt values in the range $\lambda \in [0.3, 0.7]$ consistent with prior work \cite{xu2023ipadapter}, tuning it to balance semantic coherence and visual fidelity.

This approach enables visual grounding while preserving SDXL’s pre-trained text-to-image generation capabilities. Because all original model weights (including the U-Net and text encoder) remain frozen, the adapter imposes minimal overhead and supports efficient fine-tuning. Only the projection layers for image embeddings are updated during training. The decoupled attention structure ensures that visual and textual signals are integrated without interference, which is particularly valuable in emotionally grounded tasks such as cognitive reappraisal. Here, the goal is to transform affective meaning while retaining the spatial structure and visual context of the original image—a balance that the IP-Adapter is well-suited to achieve.

\subsection{Dataset construction for stable diffusion IP-adapter}
\label{Dataset construction for stable diffusion IP-adapter}
To train the IP-Adapter module in a cognitively grounded reappraisal task, we constructed a synthetic dataset of image--prompt pairs simulating realistic affective reinterpretations. Starting from 160 base images (80 negative, 80 neutral) drawn from IAPS~\cite{lang1997international}, we used GPT-3.5~\cite{openai2023gpt35} to generate ten reappraisal prompts per image. Where available, few-shot prompting incorporated actual human reappraisals from pilot subjects to anchor generation in realistic tone and structure. Each prompt was paired with its corresponding base image and passed to Stable Diffusion XL (SDXL 1.0)~\cite{podell2023sdxlimprovinglatentdiffusion} for generation, using the IAPS image as a visual reference through the image prompt pathway.

To enhance semantic and aesthetic fidelity, we performed post-generation verification and ranking using GPT-4 Vision~\cite{openai2023gpt4vision}. For each image, GPT-4V was prompted to evaluate whether the output visually matched the intended emotional transformation described in the prompt. Images flagged as semantically inconsistent, emotionally incoherent, or containing visual artifacts were discarded. The retained images were further augmented using standard pixel-level transformations, including symmetry, cropping, noise injection, and brightness jitter, to generate ten additional variants per sample, yielding over $16000$ total image--prompt pairs. Additionally, 20\% of prompts were paraphrased using ChatGPT-3.5~\cite{openai2023gpt35} with synonym replacement strategies to introduce lexical variety and reduce overfitting.

After automated and manual quality control, including artifact filtering and sentiment consistency checks, less than 8\% of samples were removed. The resulting dataset comprised $18000$ high-quality image--prompt pairs, each reflecting a reappraised interpretation of an emotionally neutral or negative scene. This dataset was used to fine-tune the IP-Adapter while keeping all SDXL weights frozen. All prompts used in this dataset were generated directly from Whisper-transcribed human speech~\cite{radford2022robust}, and with no grammatical corrections applied, preserving natural variation and spoken syntax. 

\subsection{IP-Adapter Fine-Tuning and Semantic Guidance Scaling}
We trained the IP-Adapter module on a dataset of 18,000 paired images and captions, using the same denoising objective employed by the base diffusion model. During training, all weights of the underlying SDXL model were frozen; only the parameters of the IP-Adapter were updated. To improve generalization and enable flexible conditioning at inference, we applied classifier-free guidance dropout~\cite{ho2022classifier} randomly omitting the image embedding, the text embedding, or both with low probability. This strategy encourages the model to operate robustly under partial or missing conditioning signals. Training was conducted over 200,000 optimization steps using the AdamW optimizer (learning rate: 1e-4) with a batch size of 48, distributed across four NVIDIA L40 GPUs (48GB VRAM each). Total training time was approximately 54 hours. Representative comparisons between outputs from the pre-trained and fine-tuned models are shown in (Figure~\ref{fig:S1}), illustrating the motivation behind training the IP-Adapter module. The original Stable Diffusion model, when used without fine-tuning, frequently produced anatomically implausible outputs—such as phantom limbs, distorted body parts, or objects unnaturally misaligned with human figures. Fine-tuning the IP-Adapter substantially improved the structural coherence and semantic fidelity of the generated images, enabling more accurate and emotionally resonant visualizations of participant prompts.

At inference, the pipeline accepts both a text prompt $p$ and a reference image $I$ (or just an image if no prompt is provided). The CLIP image encoder and projection head generate a sequence of image tokens $F_i$, while the SDXL text encoder produces a sequence of text embeddings $F_t$ corresponding to $p$. The U-Net denoiser, augmented with parallel cross-attention mechanisms, synthesizes an image from noise over $T$ denoising steps, conditioned jointly on $F_t$ and $F_i$. We use the DDIM sampler~\cite{song2020denoising} with a text guidance scale of 7.5.

The relative influence of the visual reference can be modulated by scaling the image-based attention component $O_i$ with a factor $\alpha$ before it is combined with the text-based attention output $O_t$. This allows precise control over the generative balance between textual and visual information. For the Reappraisal-AI condition, $\alpha$ was dynamically adapted for each image to account for variability in semantic content across IAPS stimuli. While some images depict complex, structured scenes, others are sparse or ambiguous (e.g., involving injury or destruction), making semantic reinterpretation more difficult.


In the Describe-AI condition, $\alpha$ was fixed at 0.8 to ensure that the generated images remained closely aligned with the visual content of the original image while still integrating information from the descriptive prompt. This produced consistent generations that preserved core visual features of the input stimulus.

\subsection{Sentiment analysis of verbal reappraisals and predictive validity for affective ratings}

To evaluate the affective tone of participants' verbal responses during the task, we performed sentiment analysis directly on the subject prompts — the verbal content produced by participants in response to each image transcribed as text to input the diffusion model. We used the the multilingual RoBERTa-based model \cite{barbieri2020tweeteval} to obtain probabilities for three sentiment labels: neutral, negative and positive.  The sentiment score was computed as a continuous polarity index ranging from $-1$ (clearly negative) to $+1$ (clearly positive), incorporating the model's full label probability distribution. Each probability was weighted according to its position on the sentiment scale: $-1$ for \textit{negative}, $0$ for \textit{neutral}, and $+1$ for \textit{positive}. Formally:

\[
\text{Sentiment Score} = P(\text{negative}) \times (-1) + P(\text{neutral}) \times 0 + P(\text{positive}) \times (+1)
\]
This results in a continuous sentiment measure that reflects both the direction and strength of affective content in the subject’s prompt. This score ranges from $-1$ (clearly negative) to $+1$ (clearly positive), capturing the overall emotional polarity of the subject’s verbal response. We conducted a Pearson correlation analysis ($\rho$) to assess the relationship between the sentiment of participants’ verbal reappraisals and their self-reported affect ratings, stratified by condition and image valence. To ensure that the observed associations were not confounded by prompt length or linguistic complexity, we repeated the analysis using linear regression while controlling for word count and text complexity. The latter was quantified using the Flesch Reading Ease score (computed via the \texttt{textstat} library)~\cite{textstat2020}, where higher scores reflect simpler language.

\subsection*{Semantic alignment between verbal reappraisal and corresponding generated imagery}

To assess the semantic alignment between participants' reappraisal prompts and the content of the corresponding AI-generated images, we employed a two-step pipeline combining image captioning and embedding-based similarity analysis. For each trial in the \textit{Reappraisal AI} condition, the reappraised image was presented to GPT-4V (via the OpenAI API) along with the standardized instruction:  
\textit{"This image was generated as a positive reinterpretation of a scene. Describe what this new image communicates emotionally and semantically."}  
GPT-4V returned a free-text caption describing both the scene and its emotional message. In cases where GPT-4V declined to generate a caption—typically due to safety filters or low salience—we used a fallback captioning pipeline based on a CLIP image encoder paired with a language decoder to generate a descriptive caption.

To quantify semantic consistency, we embedded both the participant’s verbal reappraisal prompt (\textit{reappraisal prompt}) and the AI-generated image caption (\textit{image caption}) using the \texttt{all-mpnet-base-v2} model from the \texttt{sentence-transformers} library. Each text was converted into a fixed-length embedding:
\[
\mathbf{r}_i = \texttt{encode(reappraisal prompt)} \quad \quad
\mathbf{c}_i = \texttt{encode(image caption)}
\]
Semantic alignment was then computed as the cosine similarity between the two vectors:
\[
\text{Alignment}_i = \cos(\theta) = \frac{\mathbf{r}_i \cdot \mathbf{c}_i}{\|\mathbf{r}_i\| \|\mathbf{c}_i\|}
\]
Higher values indicate stronger semantic congruence between the participant’s intended reinterpretation and the emotional content inferred from the generated image. This continuous alignment metric was used in downstream correlation analyses to evaluate whether greater prompt–image coherence was predictive of reductions in negative affect and increases in perceived emotional congruence.

\section{Results}
\subsection{Stable Diffusion-generated images maintain structural fidelity in reappraisal prompting}

To systematically evaluate the effectiveness and translational potential of generative AI for cognitive emotion regulation (CER), we designed a controlled CER task based on \cite{kanske2011regulate} incorporating Stable Diffusion XL (SDXL) augmented with IP-Adapter conditioning \cite{xu2023ipadapter}. Participants were exposed to emotionally aversive stimuli drawn from IAPS, followed by a verbal reappraisal phase in which they cognitively reappraised the presented content toward a more neutral or positive interpretation. These verbal reappraisals were transcribed in real time and repurposed as conditioning prompts for the SDXL model, which generated semantically aligned visual reinterpretations grounded in both the original image and the user-supplied cognitive transformation.

\begin{figure}[h!]
  \centering
  \includegraphics[width=0.8\textwidth]{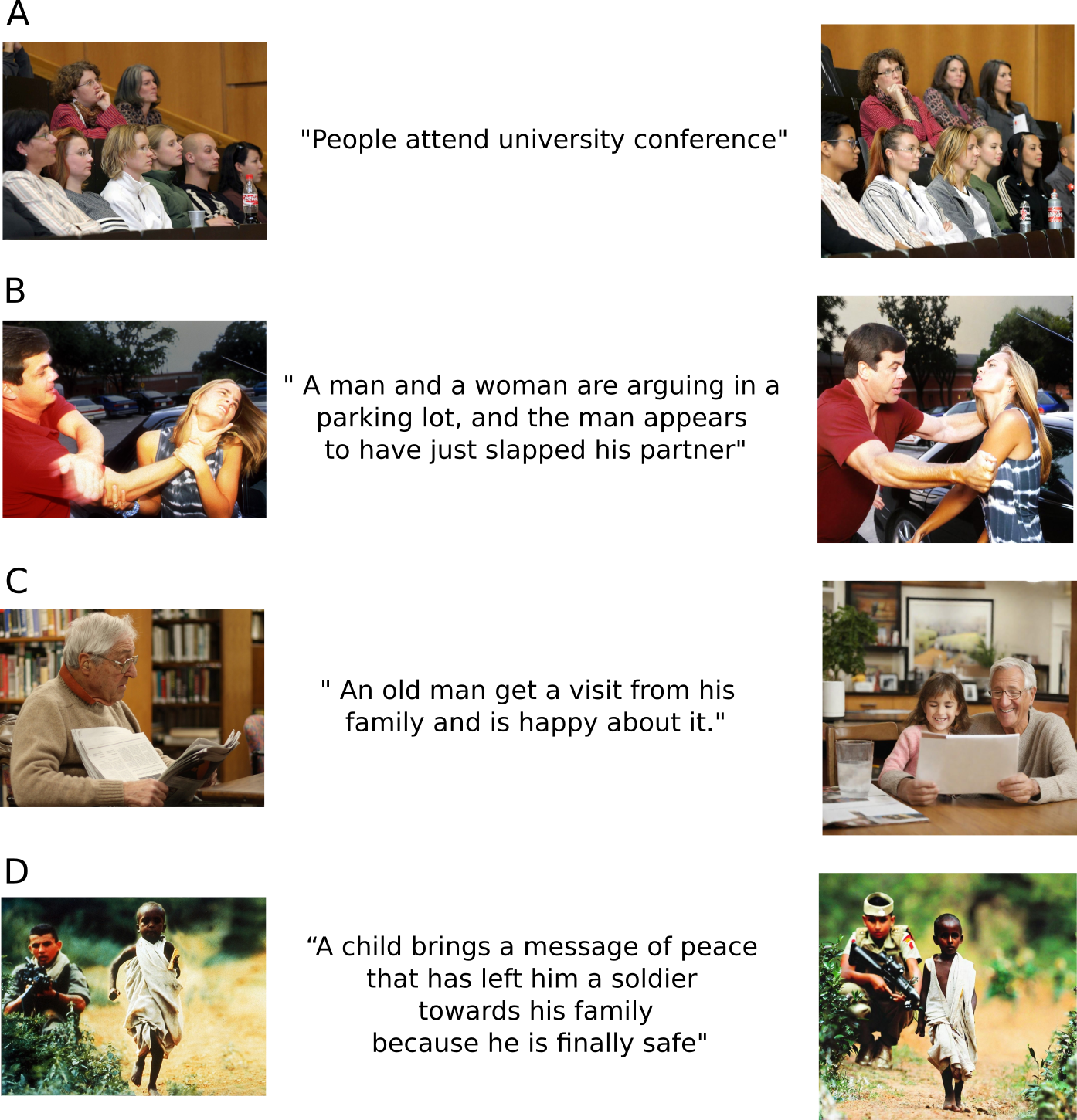}
  \caption{\textbf{Examples of reappraised image pairs generated using Stable Diffusion}. Representative examples of image transformations across experimental conditions, generated using the Stable Diffusion XL model with IP-Adapter conditioning. Each triplet shows the original IAPS stimulus (left), the transcribed verbal prompt (center), and the resulting AI-generated image (right). Prompts were captured in real time via OpenAI’s Whisper-turbo voice-to-text system and passed into the generation pipeline without manual correction, preserving participants’ spontaneous speech, including minor grammatical or typographic errors. Panels \textbf{A} and \textbf{B} illustrate outputs from the \textit{Describe-AI (neutral)} and \textit{Describe-AI (negative)} conditions, respectively. In these cases, image transformations primarily reflected the participant’s verbal descriptions of visual features from the original stimulus, preserving the original affective tone. Panels \textbf{C} and \textbf{D} show results from the \textit{Reappraisal-AI (neutral)} and \textit{Reappraisal-AI (negative)} conditions. Here, the AI-generated images exhibit clear affective transformation.
}
  \label{fig:stable_images}
\end{figure}

Figure~\ref{fig:stable_images} presents representative outputs from the generative pipeline, illustrating the effectiveness of the SDXL + IP-Adapter architecture in visually instantiating participant-generated reappraisals. Across conditions, the generated images maintained high structural fidelity to the original IAPS inputs while integrating semantic and affective cues derived from participants’ verbal prompts. In the \textit{Describe-AI (neutral)} and \textit{Describe-AI (negative)} conditions, one class of transformations involved the generation of neutral or negative image features as described by the participant based on IAPS images (Figure~\ref{fig:stable_images}A,B). In contrast, the \textit{Reappraisal-AI (neutral)} and \textit{Reappraisal-AI (negative)} conditions yielded image transformations that reflected positive reappraisals of originally neutral scenes, or
the attenuation or removal of aversive content, respectively (Figure~\ref{fig:stable_images}C,D).

It is important to note that a subset of outputs exhibited subtle artifacts or semantically diffuse renderings. These limitations are consistent with known challenges of diffusion-based models in affect-sensitive contexts, motivating further refinement of multimodal conditioning fidelity and control~\cite{blattmann2023stable,zhang2023adding,podell2023sdxlimprovinglatentdiffusion}.

\subsection{Stable Diffusion-generated images modulate emotional responses in cognitive reappraisal task}

We formally tested two hypotheses regarding the affective modulation induced by AI-augmented cognitive reappraisal. First, we hypothesized that combining cognitive reappraisal with generative visual feedback (Reappraisal AI; RAI) would significantly reduce negative affect ratings relative to traditional reappraisal without imagery (Reappraisal; R). Second, we posited that the mere inclusion of AI-generated imagery in the absence of reappraisal (Describe AI; DAI) would not significantly modulate affective ratings—either for negative or neutral stimuli—when compared to a passive description baseline (Describe), thus isolating the regulatory contribution of reappraisal semantics from the presence of imagery alone (see section ~\ref{Methods}).

As a preliminary validation of our CER paradigm, we examined the overall sensitivity of participants' affective ratings to experimental manipulation across all conditions ($N = 20$). Specifically, we compared mean emotional responses to negative and neutral images across conditions. As expected, reappraised negative images elicited significantly higher affective ratings ($M = -0.18$, $SD = 0.68$) than non-reappraised negative images ($M = -1.07$, $SD = 0.48$), with a mean difference of $0.88$. This was supported by a significant main effect of instruction, $F(1, 19) = 47.33$, $p < .001$, and a significant instruction × emotion interaction, $F(1, 19) = 15.79$, $p = .001$ (see Table~\ref{tab:anova}). These results confirm that our CER task successfully captures established emotional regulation effects, replicating patterns observed in prior literature and validating the task’s construct fidelity.


Specifically, we compared mean emotional responses to negative and neutral images across conditions. As expected, reappraised negative images elicited significantly higher affective ratings ($M = -0.18$, $SD = 0.68$) than non-reappraised negative images ($M = -1.07$, $SD = 0.48$), with a mean difference of $0.88$. This was supported by a significant main effect of instruction, $F(1, 19) = 47.33$, $p < .001$, and a significant instruction × emotion interaction, $F(1, 19) = 15.79$, $p = .001$ (see Table~\ref{tab:anova}). These results confirm that our CER task successfully captures established emotional regulation effects, replicating patterns observed in prior literature and validating the task’s construct fidelity.

\begin{table}[h!]
\centering
\begin{tabular}{lrrrr}
\toprule
index & F Value & Num DF & Den DF & Pr $>$ F \\
\midrule
emotion & 116.801100 & 1.000000 & 19.000000 & 0.000000 \\
instruction & 47.332100 & 1.000000 & 19.000000 & 0.000000 \\
modality & 25.485200 & 1.000000 & 19.000000 & 0.000100 \\
emotion:instruction & 15.790800 & 1.000000 & 19.000000 & 0.000800 \\
emotion:modality & 5.511100 & 1.000000 & 19.000000 & 0.029900 \\
instruction:modality & 28.671600 & 1.000000 & 19.000000 & 0.000000 \\
emotion:instruction:modality & 8.074900 & 1.000000 & 19.000000 & 0.010400 \\
\bottomrule
\end{tabular}
\caption{\textbf{Repeated-measures ANOVA results for emotional valence ratings}.A 2~(emotion: negative vs.\ neutral) $\times$ 2~(instruction: describe vs.\ reappraise) $\times$ 2~(modality: with AI vs.\ without AI) repeated-measures ANOVA was conducted to examine the effects of task condition on self-reported emotional valence. The table reports F-values, degrees of freedom (DF), and associated $p$-values. All main effects and higher-order interactions were statistically significant. Notably, the three-way interaction (\textit{emotion:instruction:modality}) was significant ($F(1,19) = 8.07$, $p = 0.0104$), indicating that the effectiveness of AI support varied as a function of both the emotional valence of the stimulus and the type of instruction given. This supports the interpretation that AI-generated imagery enhances reappraisal effectiveness selectively for negatively valenced stimuli. 
}
\label{tab:anova}
\end{table}



\begin{figure}[h!]
  \centering
  \includegraphics[width=0.8\textwidth]{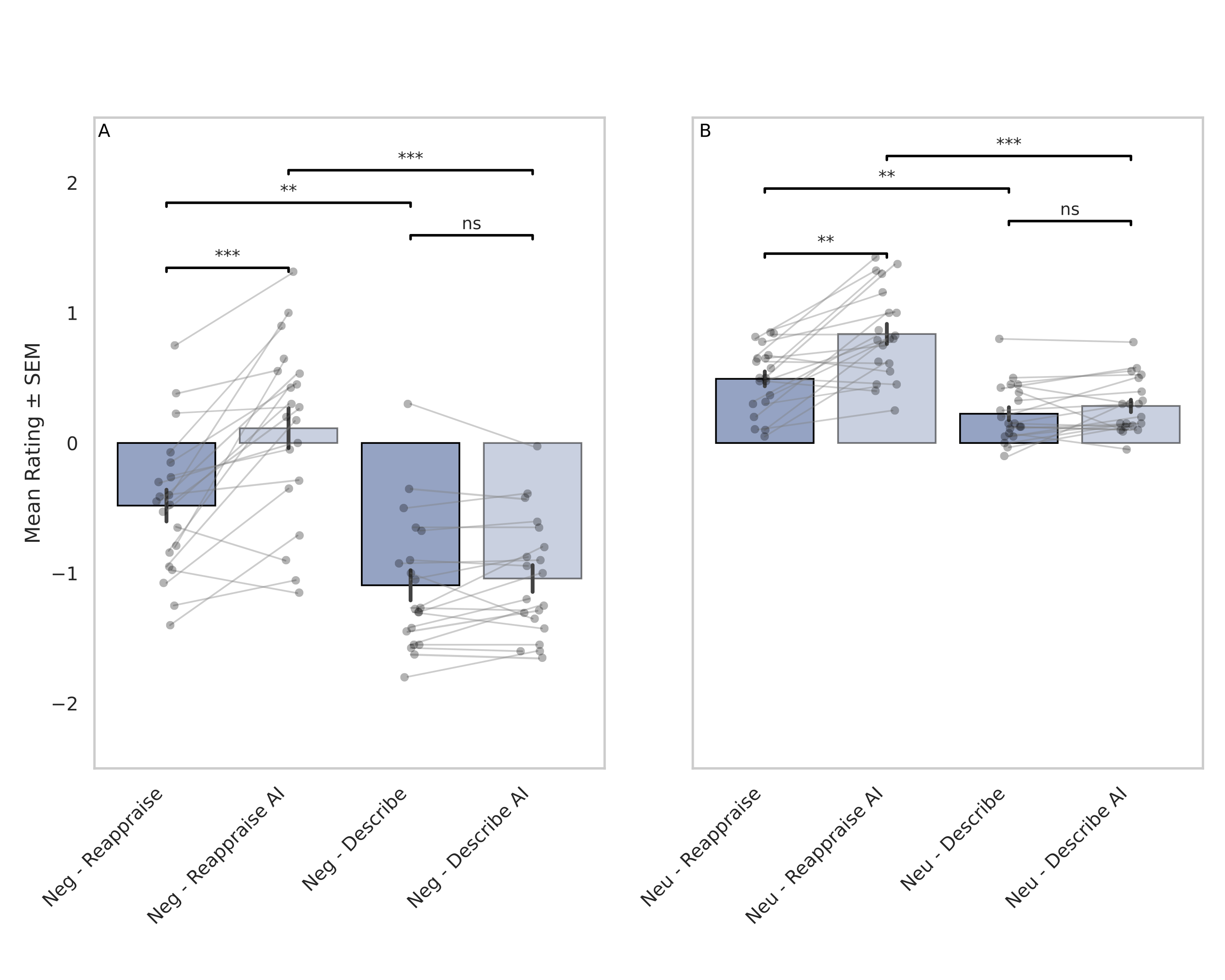}
  \caption{\textbf{Effect of AI-assisted reappraisal on emotional valence ratings across stimulus types}. Mean-centered emotional valence ratings are shown by condition for negative (\textbf{A}) and neutral (\textbf{B}) images from the IAPS dataset. Each dot represents a participant’s mean response per condition; gray lines connect repeated measures within subjects.  Statistical significance was assessed via repeated-measures ANOVA with Bonferroni-corrected pairwise comparisons. Significance markers indicate: \textbf{***}$p < 0.001$, \textbf{**}$p < 0.01$, \textbf{*}$p < 0.05$; ``ns'' denotes non-significant differences ($p \geq 0.05$). Error bars represent the standard error of the mean (SEM).}
  \label{fig:mean_ratings}
\end{figure}

To test the first hypothesis, we compared participants’ mean emotional ratings for negative images in the Reappraisal AI (Neg-RAI) and traditional Reappraisal (Neg-R) conditions ($N = 20$). For negative stimuli, participants in the Neg-RAI condition reported significantly reduced negative affect compared to those in the NEG-R condition (Figure~\ref{fig:mean_ratings}A; mean difference Neg-RAI vs Neg-R = 0.818, post-hoc t-test: \textit{t}=$5.20$, $p < .001$ (Bonferroni corrected)). Traditional reappraisal alone produced intermediate affective responses, showing improvement relative to passive descriptive viewing (Neg-D) but falling short of the benefit conferred by AI-assisted reappraisal (Figure~\ref{fig:mean_ratings}A; mean difference Neg-R vs Neg-D =  $0.6$, Post-hoc t-test: \textit{t}=$4.14$, $p < .05$, Bonferroni corrected). 

Along those lines, affective ratings of negative images in the Reappraisal AI (Neg-RAI) were significantly lower than the descriptive condition with AI-generated imagery (Neg-DAI; mean difference Neg-RAI vs Neg-DAI = $1.15$; post-hoc t-test: \textit{t}=$7.107$, $p < .001$, Bonferroni corrected)). Taken together, these results provide strong empirical support for the first hypothesis, underscoring the additive regulatory value of AI-generated visual reinterpretation in amplifying the cognitive mechanisms of reappraisal.

To evaluate the second hypothesis, we assessed whether affective ratings for the descriptive viewing of negative images (Neg-D) differed from those in the descriptive condition with AI-generated imagery (Neg-DAI). As shown in Figure~\ref{fig:mean_ratings}A, there was no significant difference between these conditions (mean difference = $0.05$, n.s.), indicating that the observed AI-driven effects were specific to the reappraisal of emotionally aversive content rather than to its mere verbal description or visual augmentation.

We further examined whether affective ratings differed between the Neu-D and Neu-DAI conditions, which involved the descriptive viewing of neutral images with and without AI-generated imagery. Consistent with Hypothesis 2, no statistically significant difference was observed between these conditions (mean difference = $0.06$, n.s.; Figure~\ref{fig:mean_ratings}B). This finding reinforces the interpretation that the presence of AI-generated imagery alone is insufficient to modulate affect in the absence of explicit cognitive reappraisal. These null results support the conclusion that the affective improvements observed in the Neg-RAI condition (negative Reappraisal with AI) are not attributable to visual input alone, but rather to the interaction between semantic reinterpretation and regulatory intent embedded in the reappraisal process.

Interestingly, we also tested whether AI-assisted reappraisal could enhance affective modulation for neutral stimuli (Figure~\ref{fig:mean_ratings}B). Here, a significant difference emerged: the neutral Reappraisal AI (Neu-RAI) condition yielded significantly higher (i.e., more positive) ratings than the traditional reappraisal (Neu-R) condition (mean difference = $0.34$, post-hoc t-test: \textit{t}=$4.77$, $p < .01$, Bonferroni corrected)). This finding suggests that even in low-arousal contexts, coupling cognitive reinterpretation with personalized generative imagery amplifies the emotional regulatory effect, further supporting the synergistic benefit of combining reappraisal content with visual scaffolding.

Taken together, across both negative and neutral conditions, no statistically significant differences were observed between the Describe AI and Describe-only groups. This reinforces the interpretation that, in the absence of explicit cognitive reappraisal, AI-generated images alone do not meaningfully alter emotional experience (see Figure~\ref{fig:mean_ratings}A-B). These findings further underscore the specificity of the cognitive-AI interaction and highlight the necessity of coupling generative output with intentional, user-driven reinterpretation to achieve effective emotion regulation.

\subsection{Verbal reappraisal sentiment correlates with affective outcome}

\begin{figure}[h!]
  \centering
  \includegraphics[width=0.6\textwidth]{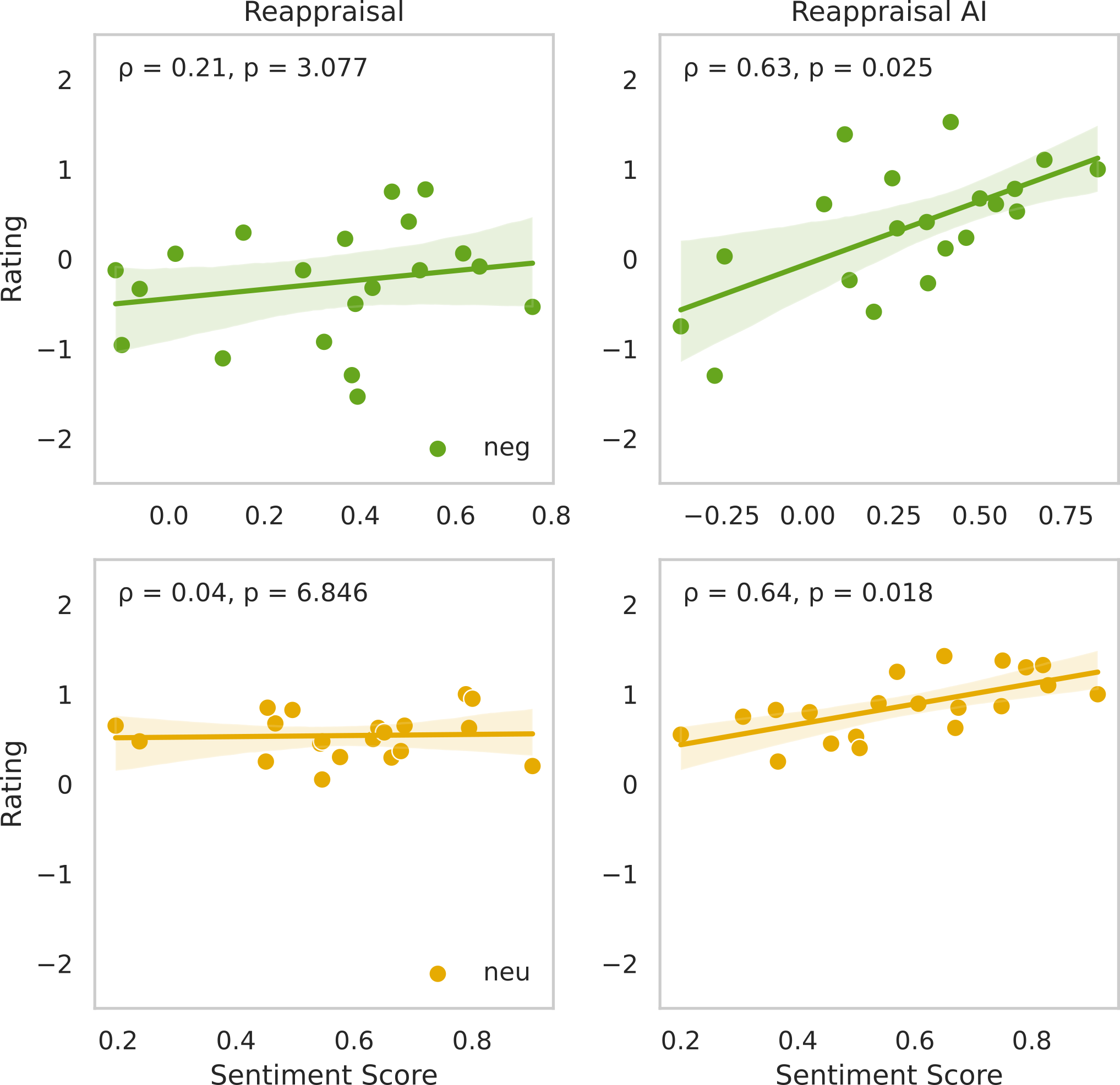}
  \caption{\textbf{Correlation between sentiment of reappraisal prompts and affective ratings across task conditions.} Scatter plots show the relationship between mean sentiment scores (x-axis) and participant-reported affective mean ratings (y-axis) under two task conditions (columns: Reappraisal, Reappraisal AI) and two stimulus types (rows: negative stimuli, top, green; neutral stimuli, bottom yellow). Mean values were computed by averaging across all trials under each condition and stimulus type. Each subplot reports Pearson correlation coefficient ($\rho$) and the associated two-tailed Bonferroni corrected $p$-value. Shaded areas around regression lines indicate standard error of the mean (SEM), reflecting uncertainty in the fit across subjects.}
  \label{fig:correlation}
\end{figure}

Next, we wished to investigate the relationship between reappraisal content and emotional outcomes, we analyzed the correlation between sentiment scores extracted from participants’ verbal reappraisal prompts and their corresponding affect ratings. Verbal inputs were transcribed using automatic speech-to-text and analyzed using a standard sentiment analysis model, which produced continuous scores reflecting the emotional valence of each utterance, ranging from –1 (strongly negative) to +1 (strongly positive; see~\ref{Methods}). To evaluate whether the affective tone of participants’ verbal prompts correlated with  their emotional ratings, we conducted correlations analysis between the sentiment score of each reappraisal or description prompt and the corresponding affective rating, separately for each experimental condition and stimulus valence (Figure~\ref{fig:correlation}).

For negative-valence trials, a significant positive correlation was observed in the \textit{Reappraisal-AI (Neg-RAI)} condition ($\rho = 0.63$, $p = 0.025$), indicating that more positively valenced reappraisal prompts were associated with more favorable (less negative) affective ratings. This relationship was absent in the traditional reappraisal condition (\textit{Neg-R}: $\rho = 0.21$, $p = 0.3077$), suggesting that the verbal generation of positive reinterpretations alone does not consistently translate into improved emotional outcomes. Instead, these findings support the hypothesis that AI-generated visual imagery plays a critical role in reinforcing the emotional intent of reappraisal content, enabling more effective affective modulation through multimodal alignment.

For neutral-valence trials, a similar pattern emerged. The \textit{Reappraisal-AI (Neu-RAI)} condition showed a statistically significant correlation between prompt sentiment and affective ratings ($\rho = 0.64$, $p = 0.018$), whereas no such association was observed in the \textit{Reappraisal (Neu-R)} condition ($\rho = 0.04$, $p =6.8 0.49$). This suggests that even in lower-arousal contexts, the presence of visual scaffolding enhances the efficacy of verbal reinterpretations by grounding them in emotionally congruent imagery.

Together, these results underscore the importance of semantic and emotional alignment between participants' reappraisals and the corresponding AI-generated visual outputs \cite{sharma2023human, glickman2025human}. While sentiment alone is insufficient to drive affective change—as evidenced by the weak or absent correlations in the no-AI conditions—its predictive power emerges when reinforced through personalized generative imagery. These findings support our central hypothesis: that generative AI can serve as an effective amplifier of cognitive reappraisal by transforming abstract regulatory intent into emotionally resonant visual content.


\subsection{Semantic alignment between reappraisal prompts and generated images predicts affective outcome}

To assess whether the effectiveness of AI-assisted reappraisal depends on the degree of semantic congruence between participants’ verbal input and the resulting AI-generated imagery, we quantified alignment scores (range from –1, completely dissimilar, to +1, perfectly similar) between each participant's spoken reappraisal prompt and the caption of the corresponding generated image. These captions were derived using a vision-language model (GPT-4V or a fallback CLIP-based captioning model; see~\ref{Methods}), and both captions and prompts were embedded using a sentence transformer to compute cosine similarity.

\begin{figure}[h!]
  \centering
  \includegraphics[width=0.8\textwidth]{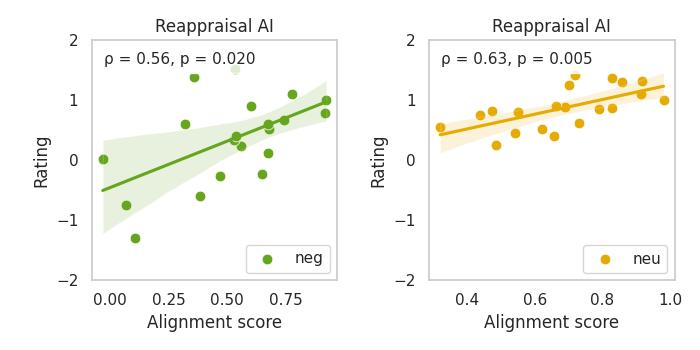}
  \caption{\textbf{Correlation between image–prompt alignment and affective ratings in the Reappraisal AI condition}. Scatter plots show the relationship between alignment scores (x-axis) and participant-reported emotional valence ratings (y-axis) for negative (left, green) and neutral (right, yellow) stimuli, both within the Reappraisal AI condition. Alignment scores quantify the semantic similarity between participant-generated reappraisal prompts and generated captions of the corresponding AI-generated images, computed via cosine similarity of sentence embeddings (using Sentence-BERT \cite{reimers2019sentence}). Each subplot reports Pearson correlation coefficient ($\rho$) and associated two-tailed Bonferroni corrected $p$-value. Shaded areas indicate the standard error of the mean (SEM) for the regression fit.}
  \label{fig:multimodal_align}
\end{figure}

As shown in Figure~\ref{fig:multimodal_align}, semantic alignment between the reappraisal prompt and the generated image was significantly correlated with participants’ affective ratings. This effect was evident for both negative and neutral image contexts. In the Reappraisal-AI (Neg-RAI) condition, alignment scores were positively associated with less negative ratings ($\rho = 0.56$, $p = 0.020$), suggesting that the closer the visual output mirrored the user’s reappraisal intent, the greater the emotional benefit. Similarly, in the Reappraisal-AI (Neu-RAI) condition, a significant correlation was observed ($\rho = 0.63$, $p = 0.005$), indicating that even in low-arousal contexts, better alignment enhances the regulatory outcome.

These results suggest that the efficacy of AI-assisted emotion regulation is modulated by multimodal coherence—that is, the extent to which the emotional and semantic content of the generated image faithfully reflects the user’s intended reinterpretation. Together, the findings provide further evidence that generative models can amplify cognitive reappraisal when their outputs are tightly aligned with user-driven affective goals.

\section{Discussion}

This study demonstrates that integrating generative AI imagery with cognitive reappraisal significantly enhances affective regulation compared to traditional reappraisal alone. Specifically, participants in the Reappraisal AI (RAI) condition reported more positive emotional outcomes than those in the Reappraisal (R) condition, supporting our first hypothesis. In contrast, emotional ratings in the Describe AI (DAI) condition were indistinguishable from those in the Describe (D) condition across both negative and neutral stimuli, strongly supporting our second hypothesis and reinforcing that the observed affective benefits were driven by reappraisal content rather than the mere presence of AI-generated images. Additionally, sentiment analysis revealed a strong correlation between the positivity of verbal reappraisal prompts and subsequent emotional ratings, while semantic alignment between participant prompts and generated images significantly predicted regulatory success. Together, these findings highlight the potential of generative AI not just as a creative tool, but as a real-time cognitive scaffold capable of augmenting human emotion regulation.

Our findings advance our understanding of cognitive emotion regulation by demonstrating that generative AI can function as an effective augmentative scaffold when tightly coupled with user-driven reappraisal. From a cognitive neuroscience perspective, this work extends existing models of reappraisal by embedding them in a multimodal, interactive loop—bridging internal verbal reframing with externalized visual feedback \cite{ruppert2018positive, yamada2018exploratory, pearson2015mental}. The emotional gains observed in the Reappraisal AI (RAI) condition, relative to traditional reappraisal, suggest that pairing generative models with human cognitive processes can amplify the regulatory effect through richer semantic encoding and reinforcement. Crucially, the absence of affective change in the Describe AI (DAI) condition underscores that the benefits are not attributable to the imagery alone but emerge from the semantic and intentional structure of reappraisal itself.

In the context of affective computing and human-AI collaboration, our results point to a novel use case for generative models: as emotionally aligned, user-conditioned agents capable of participating in shared cognitive tasks \cite{sharma2023human}. Prior systems in this space—such as LLM-based chatbots or unimodal text reappraisal tools—have largely treated emotional support as a conversational or informational problem \cite{bell2024advances,blackwell2021mental,prochaska2021therapeutic, vaidyam2019chatbots}. In contrast, our approach treats reappraisal as a co-regulated perceptual process, where the AI system visually instantiates the user’s intent in real time. This opens pathways toward emotion-aware, interactive systems that respond not just to what users feel, but to how they think about what they feel \cite{rubin2025comparing, glickman2025human}.

From a translational perspective, the integration of generative AI into therapeutic contexts such as cognitive behavioral therapy (CBT) represents a promising frontier. Our paradigm demonstrates that personalized, semantically grounded visual feedback can enhance emotional self-regulation in a controlled setting, suggesting potential for deployment in digital mental health platforms \cite{bucci2019digital,dehbozorgi2025application}. Future tools could incorporate generative visual models as adaptive, real-time assistants—augmenting reappraisal strategies, increasing emotional engagement, and lowering cognitive load for users with stress-related or mood disorders \cite{miner2017talking,lincoln2022role}. The paradigm may also hold particular value for populations with limited verbal fluency or executive function, by externalizing abstract thought into concrete visual form \cite{ruppert2018positive,yamada2018exploratory,pearson2015mental}.

\subsection{Limitations of this study}

While the present findings offer compelling evidence for the efficacy of AI-assisted reappraisal, several limitations warrant consideration. First, the sample size ($N = 20$) and composition limits statistical power and generalizability, particularly across demographic, cultural, and clinical subpopulations. Second, although sentiment and semantic alignment metrics provided interpretable proxies for affective content, the underlying models (e.g., sentiment analyzers and image captioning systems) are themselves imperfect and may introduce noise or cultural bias \cite{luccioni2023stable}. Third, the study relied exclusively on self-reported affect ratings, which—while widely used in affective science—lack the granularity of physiological or behavioral measures \cite{korpal2018physiological}.

In terms of technical constraints, the generative model occasionally produced visually ambiguous or abstract imagery, raising questions about interpretability and user trust in emotionally sensitive contexts. Furthermore, because the IAPS dataset \cite{lang1997international} was not designed for use with diffusion-based generative models, this may have constrained the semantic fidelity of generated outputs. We also limited the SDXL inference process to 40 denoising steps to maintain real-time responsiveness, which may have impacted image quality and emotional clarity \cite{xu2023ipadapter}. We also introduced a 4-second silent delay after participants provided their verbal prompt to allow for generation latency. However, this temporal gap may have weakened the immediacy of emotional engagement with the task, potentially reducing the intensity of affective experience and making it harder to detect regulatory effects. Future designs might benefit from tighter stimulus-response coupling or immersive generation environments.

Beyond technical constraints, there are also conceptual and methodological considerations. Although the Describe condition served as a control for verbal processing, prior work suggests that verbal description alone can elicit automatic forms of emotion regulation \cite{powers2019regulating, nook2022linguistic, lieberman2007putting}. This may have reduced the contrast between Describe and Reappraisal conditions, potentially underestimating the unique contribution of reappraisal. Nevertheless, we still observed a significant effect of reappraisal with AI support, indicating that the intervention has added value even against this active baseline.

Moreover, while the Describe AI condition was not central to our primary hypotheses, post hoc analyses revealed that alignment and sentiment scores were significantly correlated with affective ratings for negative stimuli (Figure~\ref{fig:S2},~\ref{fig:S3}) . On the surface, this may suggest that AI-generated imagery can influence emotional responses even in the absence of explicit reappraisal intent. However, a closer inspection reveals a crucial distinction: although alignment improved affect ratings in Describe AI, participants’ emotional responses remained negative on average. In contrast, the Reappraisal AI condition not only showed strong alignment–rating correlations but also produced a consistent shift toward positive affect. This divergence suggests that visual-semantic congruence alone is insufficient for successful emotion regulation unless it is grounded in intentional cognitive reframing. In other words, AI-generated imagery appears to amplify the user’s regulatory intent rather than act as a standalone intervention. This interpretation reinforces our central claim that the efficacy of Reappraisal AI is driven not merely by alignment or sentiment, but by the synergistic interaction between generative visual content and reappraisal-oriented verbal input. Nonetheless, future work should more systematically disentangle the unique and interactive effects of alignment, affective tone, and reappraisal intent to clarify the mechanisms underlying these observed benefits.

Finally, our findings reflect short-term emotional modulation within a controlled experimental setting. Future work is needed to evaluate the durability of these effects, their generalizability to naturalistic environments, and the broader ethical implications of using generative models in real-time emotional interventions \cite{rubin2025comparing, pataranutaporn2023influencing, sharma2023human, glickman2025human}.

\subsection{Conclusion}

We presented a multimodal framework that integrates diffusion-based generative models with user-driven cognitive reappraisal to support emotion regulation. By conditioning image generation on verbal reinterpretations of emotionally negative stimuli, the system produces personalized visual outputs that reflect user input while preserving the structural content of the original image. Our results show that AI-assisted reappraisal leads to greater reductions in negative affect compared to traditional reappraisal, and that neither AI-generated imagery nor verbal description alone modulates affective ratings. Further, alignment between user intent and generated image content was predictive of regulatory success, suggesting that semantic consistency plays a critical role in outcome efficacy. While our findings are limited to short-term, self-reported measures, they highlight the feasibility of combining generative models with cognitive scaffolding tasks. Future work should explore how such systems can be extended, personalized, and safely deployed in real-world affective computing and mental health applications.

\section{Acknowledgments}
This work was supported by Daimler Benz foundation grant (A.F.C.), and by LIR and MLU core funding (E.P. and O.T.). We thank our colleagues for insightful discussions and feedback throughout the project. Finally, we would like to thank Georgios Michalareas for support setting the behavioral experiments.

\section*{Author Contributions}

\textbf{E.P.} contributed to the conceptualization, methodology, data curation, software development, formal analysis, investigation, writing of the original draft, review and editing of the manuscript, and visualization. \textbf{O.T.} contributed to the conceptualization, methodology, writing of the original draft, review and editing of the manuscript, and funding acquisition. \textbf{A.F.C.} contributed to the conceptualization, methodology, data curation, investigation, writing of the original draft, review and editing of the manuscript, visualization, and funding acquisition.

\bibliographystyle{IEEEtran}
\bibliography{sample.bib}

\clearpage
\section{Supplementary figures}
\setcounter{figure}{0} 
\renewcommand{\thefigure}{S\arabic{figure}} 
\begin{figure}[h!]
  \centering
  \includegraphics[width=0.6\textwidth]{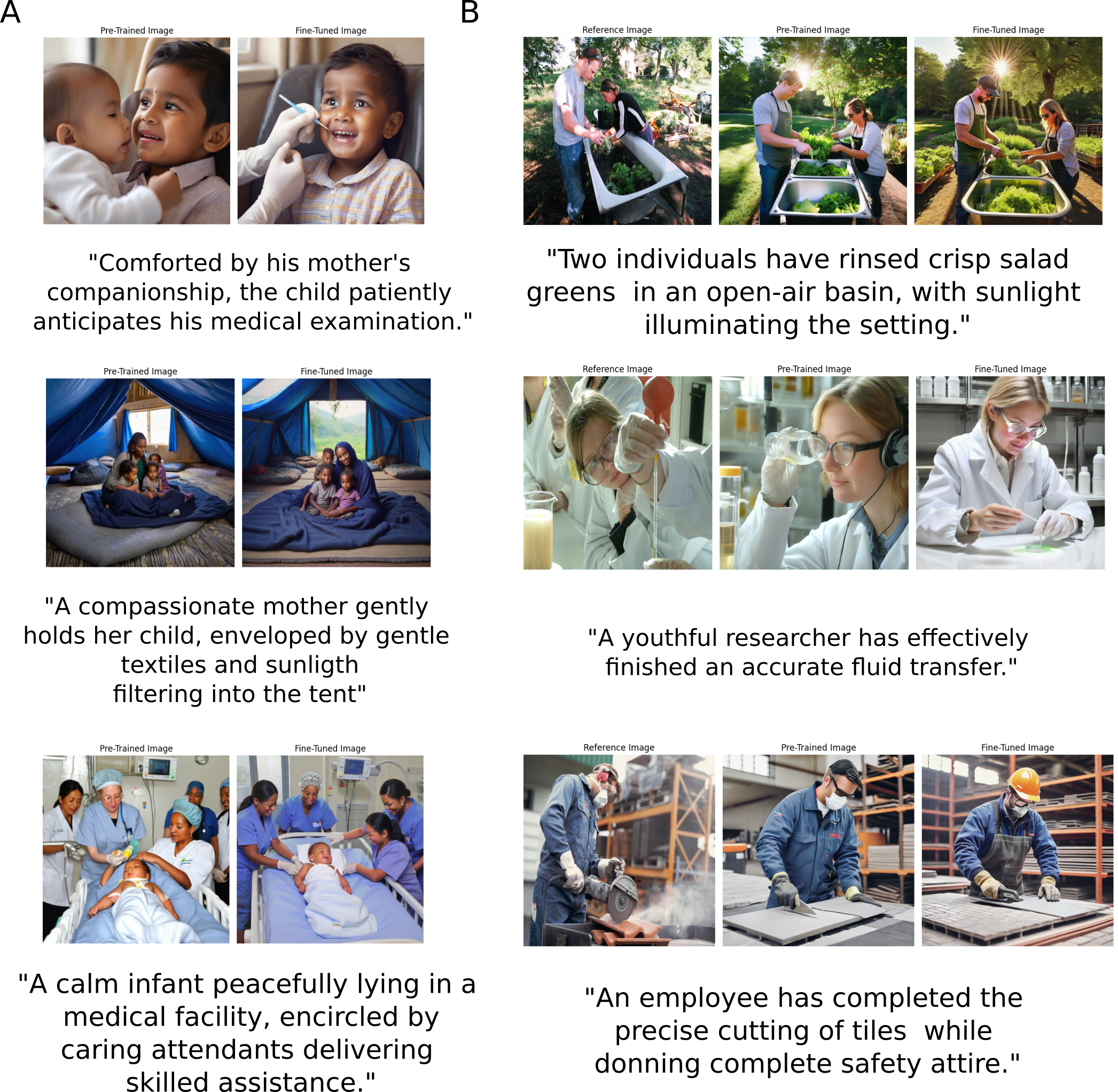}
  \caption{\textbf{Visual Comparison of pre-trained and fine-tuned outputs from the SDXL + IP-Adapter Model.} (\textbf{A}). Examples of emotionally positive reappraisals generated before and after 200,000 steps of fine-tuning on paired reappraisal image–prompt data. Prompts were held constant, while outputs reflect the difference between the base (pre-trained) and fine-tuned SDXL-IP-Adapter model. Reference IAPS images are omitted due to potential distressing content. Notably, the pre-trained model often failed to render anatomically coherent scenes: in the top row, an additional baby was erroneously generated beside the primary child, creating an unnatural duplication. In the middle row, the central child appears to have a phantom limb; and in the bottom row, one medical professional is shown holding an extra, disembodied arm—an artifact not intended in the prompt. In contrast, the fine-tuned model produces scenes that are anatomically plausible, semantically aligned with the prompt, and emotionally coherent. (\textbf{B}). Representative examples of neutral-valence image prompts alongside outputs at three stages: the original reference image (left), pre-trained model output (middle), and fine-tuned model output (right). In the top row, both the pre-trained and fine-tuned outputs show reasonable semantic fidelity; however, the fine-tuned version offers greater compositional structure and detail. In the middle row, the pre-trained model inaccurately fused a lab beaker to the subject’s glasses—likely due to visual ambiguity in the reference—whereas the fine-tuned model correctly depicts a fluid transfer task. In the bottom row, the pre-trained model produces a worker with facial features resembling a synthetic wig rather than appropriate safety gear, while the fine-tuned model correctly renders a hard hat and protective equipment in line with the original prompt. Together, these examples illustrate how fine-tuning substantially improves anatomical accuracy, prompt-image coherence, and emotional plausibility—key elements for successful emotion regulation support.
}
\label{fig:S1}
\end{figure}

\clearpage
\begin{figure}[h!]
  \centering
  \includegraphics[width=0.6\textwidth]{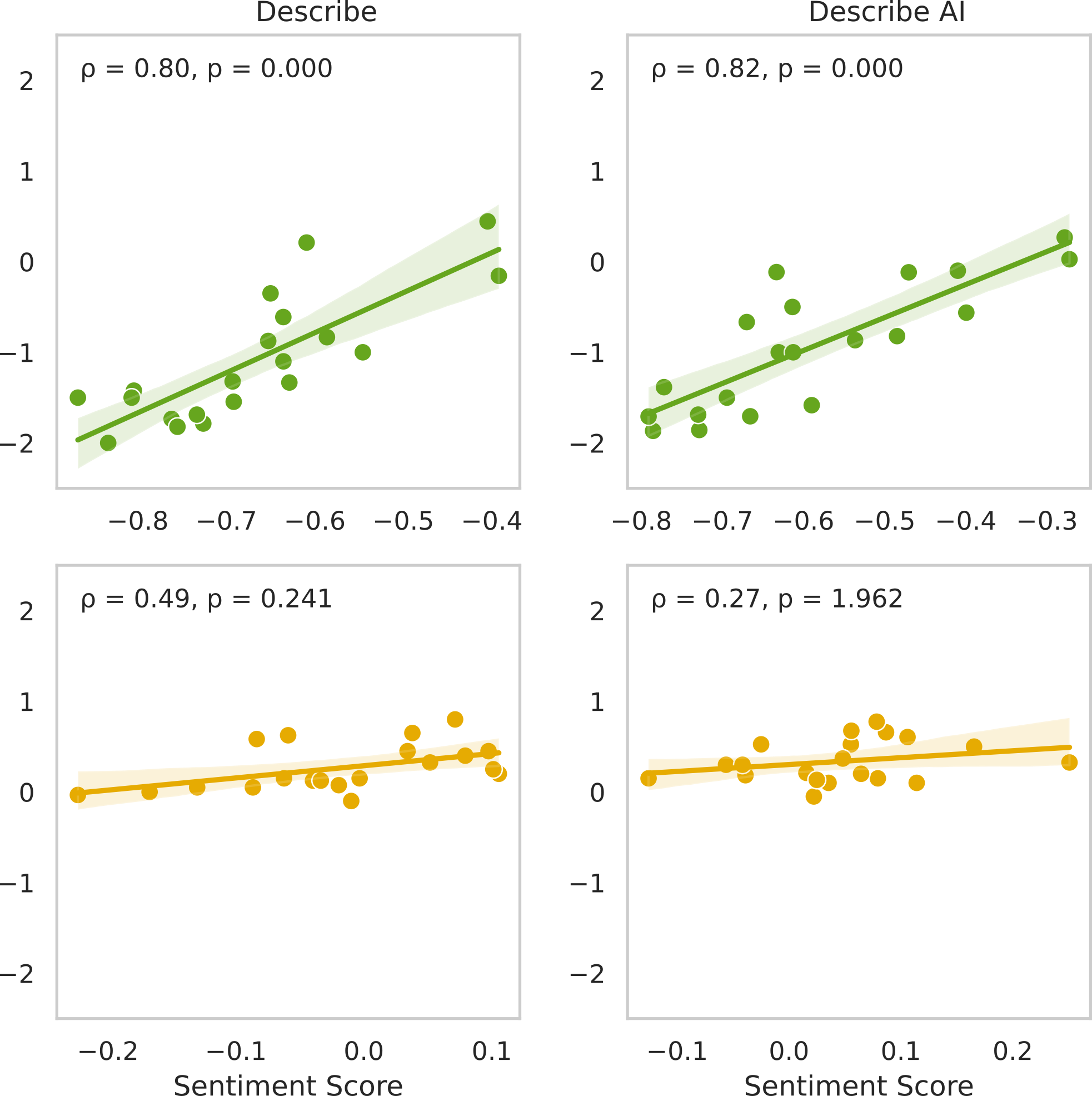}
  \caption{\textbf{Correlation between sentiment of reappraisal prompts and affective ratings across Describe and Describe AI conditions.} Scatter plots show the relationship between mean sentiment scores (x-axis) and participant-reported affective mean ratings (y-axis) under two task conditions (columns: Describe, Describe AI) and two stimulus types (rows: negative stimuli, top, green; neutral stimuli, bottom yellow). Mean values were computed by averaging across all trials under each condition and stimulus type. Each subplot reports Pearson correlation coefficient ($\rho$) and the associated two-tailed $p$-value. Shaded areas around regression lines indicate standard error of the mean (SEM), reflecting uncertainty in the fit across subjects.}
  \label{fig:S2}
\end{figure}

\clearpage
\begin{figure}[h!]
  \centering
  \includegraphics[width=0.8\textwidth]{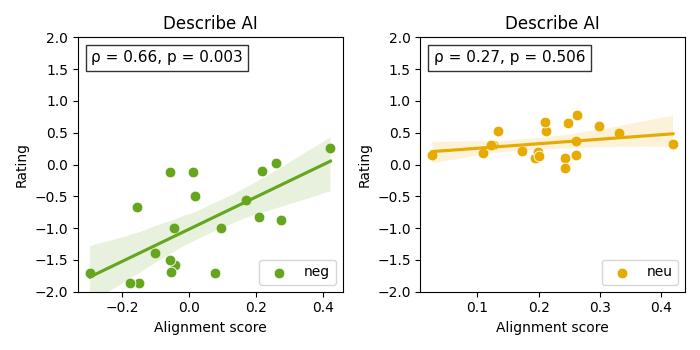}
  \caption{\textbf{Correlation between image–prompt alignment and affective ratings in the Describe AI condition}. Scatter plots show the relationship between alignment scores (x-axis) and participant-reported emotional valence ratings (y-axis) for negative (left, green) and neutral (right, yellow) stimuli, both within the Reappraisal AI condition. Alignment scores quantify the semantic similarity between participant-generated reappraisal prompts and generated captions of the corresponding AI-generated images, computed via cosine similarity of sentence embeddings (using Sentence-BERT \cite{reimers2019sentence}). Each subplot reports Pearson correlation coefficient ($\rho$) and associated two-tailed $p$-value. Shaded areas indicate the standard error of the mean (SEM) for the regression fit.}
  \label{fig:S3}
\end{figure}

\end{document}